\documentclass{ws-procs11x85}
\usepackage{ws-procs-thm}           

\usepackage{hyperref}
\usepackage{booktabs}
\usepackage{array}
\usepackage{multirow}

\let\oldcaption\caption 
\let\oldcaptionbox\captionbox
\let\caption\relax 
\let\captionbox\relax
\usepackage{caption} 
\let\caption\oldcaption 
\let\captionbox\oldcaptionbox

\begin{document}

\title{Prototype Learning to Create Refined Interpretable Digital Phenotypes from ECGs}


\author{Sahil Sethi\textsuperscript{1,2,*}, David Chen\textsuperscript{2,*}, Michael C. Burkhart\textsuperscript{2} , Nipun Bhandari\textsuperscript{3} , Bashar Ramadan\textsuperscript{4}, and Brett Beaulieu-Jones\textsuperscript{2,$\dag$}}

\address{\textsuperscript{1}\textit{Pritzker School of Medicine, University of Chicago, IL, USA}} 

\address{\textsuperscript{2}\textit{Center for Computational Medicine \& Clinical AI, Section of Biomedical Data Science, Department of Medicine, University of Chicago, IL, USA}}

\address{\textsuperscript{3}\textit{Division of Cardiovascular Medicine, Department of Internal Medicine, University of California Davis, CA, USA}}

\address{\textsuperscript{4}\textit{Section of Hospital Medicine, Department of Medicine, University of Chicago, IL, USA}}

$^\dag$E-mail: beaulieujones@uchicago.edu, \textit{* authors contributed equally}

\begin{abstract}
Prototype-based neural networks offer interpretable predictions by comparing inputs to learned, representative signal patterns anchored in training data. While such models have shown promise in the classification of physiological data, it remains unclear whether their prototypes capture an underlying structure that aligns with broader clinical phenotypes. We use a prototype-based deep learning model trained for multi-label ECG classification using the PTB-XL dataset. Then without modification we performed inference on the MIMIC-IV clinical database. We assess whether individual prototypes, trained solely for classification, are associated with hospital discharge diagnoses in the form of phecodes in this external population. Individual prototypes demonstrate significantly stronger and more specific associations with clinical outcomes compared to the classifier's class predictions, NLP-extracted concepts, or broader prototype classes across all phecode categories. Prototype classes with mixed significance patterns exhibit significantly greater intra-class distances (p $<$ 0.0001), indicating the model learned to differentiate clinically meaningful variations within diagnostic categories. The prototypes achieve strong predictive performance across diverse conditions, with AUCs ranging as high as 0.89 for atrial fibrillation to 0.91 for heart failure, while also showing substantial signal for non-cardiac conditions such as sepsis and renal disease. These findings suggest that prototype-based models can support interpretable digital phenotyping from physiologic time-series data, providing transferable intermediate phenotypes that capture clinically meaningful physiologic signatures beyond their original training objectives.

\end{abstract}


\copyrightinfo{\copyright\ 2025 The Authors. Open Access chapter published by World Scientific Publishing Company and distributed under the terms of the Creative Commons Attribution Non-Commercial (CC BY-NC) 4.0 License.}

\section{Introduction}\label{aba:intro}
Modern healthcare systems generate vast quantities of physiologic time-series data, yet our ability to extract clinically meaningful structure from these signals remains limited. Electrocardiograms (ECGs), in particular, encode rich information about cardiac and systemic physiology. While deep learning models have demonstrated impressive accuracy in ECG-based diagnosis \cite{hannun_cardiologist-level_2019, elias_deep_2022}, their opaque representations offer little insight into how physiologic patterns relate to broader patient phenotypes. Recently, Hughes et al.\cite{hughes_deep_2025} demonstrated that ECGs can detect a surprisingly broad spectrum of 1,243 different cardiac and non-cardiac conditions, including previously unknown phenotypes such as neutropenia and menstrual disorders, while revealing that many non-cardiac conditions share similar ECG signatures. 

As precision medicine aims to stratify patients based on biologically meaningful traits, there is growing interest in models that not only predict, but also explain and discover. Prototype-based neural networks offer an interpretable alternative to black-box classifiers by grounding predictions in similarity to representative signal patterns \cite{chen_this_2019, barnett_case-based_2021, barnett_improving_2024, sethi_protoecgnet_2025}. Recent work has shown that such models can yield clinically coherent explanations in ECG classification tasks \cite{sethi_protoecgnet_2025}, suggesting a promising foundation for data-driven phenotyping.

In this work, we explore whether the latent structure learned by a prototype-based ECG model can reveal informative associations with real-world clinical phenotypes. Rather than training a model for diagnostic prediction, we use a prototype-based model to perform inference on ECGs from the MIMIC-IV database \cite{johnson_mimic-iv_2023}, and assess whether specific waveform prototypes correspond to structured phenotypes such as discharge diagnoses. \emph{Our central hypothesis is that learned ECG prototypes—despite being trained for a different task—may capture transferable physiologic signatures linked to disease.}

This study bridges physiologic modeling and clinical phenotyping by demonstrating how interpretable, prototype-based representations learned for the classification of raw ECGs can be associated with both cardiac and unrelated phenotypes. This reframes the potential of prototype-based models not just as diagnostic tools, but as interpretable instruments with the potential to reveal clinically meaningful structure in physiologic signals.

\section{Related Work}
\subsection{Prototype-Based Learning for Interpretability}
Prototype-based neural networks offer a transparent alternative to black-box deep learning and post-hoc methods by grounding predictions in similarity to a learned set of representative patterns \cite{chen_this_2019}. Originally developed for image classification, these models have been extended to time-series domains such as electroencephalography (EEG) and ECG, where each prototype is anchored to a localized segment of the input signal \cite{sethi_protoecgnet_2025, barnett_improving_2024}. Recent work has demonstrated that prototype-based ECG models can produce clinically meaningful explanations and achieve competitive performance on multi-label tasks \cite{sethi_protoecgnet_2025}. However, prior applications focus on supervised classification, and to our knowledge, none have explored whether learned prototypes are associated with structured clinical outcomes.  

\subsection{Phenotyping from Physiologic Signals}

Computational phenotyping plays a central role in precision medicine by enabling data-driven identification of patient subgroups that cut across conventional diagnostic boundaries. Existing phenotyping frameworks often rely on structured electronic health record (EHR) data such as ICD codes, lab results, and medication history \cite{hripcsak_next-generation_2013, denny_systematic_2013}, with limited incorporation of raw physiologic signals. However, with recent advances in deep learning, there has been a growing interest in using electrocardiograms (ECGs) beyond conventional cardiac classification tasks for phenotype discovery. Recent large-scale studies have demonstrated that ECGs encode features associated with a wide range of cardiac and non-cardiac phenotypes, including heart failure, chronic kidney disease, and sepsis \cite{hughes_deep_2025, friedman_unsupervised_2025}. Friedman et al.\ used unsupervised ECG autoencoders to derive latent representations linked to structured outcomes through phenome-wide association studies (PheWAS) across multiple cohorts \cite{friedman_unsupervised_2025}. Similarly, PheWASNet demonstrated that supervised deep learning models trained on raw ECGs can predict over 1000 phenotypes mapped from EHRs—though without interpretability \cite{hughes_deep_2025}. An important difference is that their approach set out specifically to map ECGs to phenotypes. While these models highlight the utility of ECG signals, their latent embeddings are difficult to interpret and offer limited insight into which waveform characteristics contribute to prediction. This work learns interpretable features from ECGs by performing a different task and then examines whether these features are associated and predictive of both related and unrelated phenotypes.

\section{Methods}
\subsection{Data}
\subsubsection{PTB-XL Dataset}
The publicly available PTB-XL dataset \cite{wagner_ptb-xl_2020} was used for model training. PTB-XL contains 21,837 12-lead ECG recordings from 18,885 patients, each sampled at 100~Hz over 10 seconds. Each ECG is annotated with one or more diagnostic statements, mapped to the SCP coding scheme, spanning rhythm disturbances, morphological abnormalities, and conduction disorders. We adopted the 100~Hz version of the signals and followed a multi-label classification setup with all 71 labeled classes. The original stratified 8:1:1 split from the dataset was used for training, validation, and test sets.
\subsubsection{MIMIC-IV ECG Subset}
For downstream phenotyping, we used 12-lead ECG recordings from the MIMIC-IV database \cite{johnson_mimic-iv_2023}. ECGs were resampled to 100~Hz. MIMIC-IV does not contain expert ECG annotations, so structured metadata on hospital discharge diagnoses (ICD-9 and ICD-10) was used for the downstream association analyses. Diagnoses codes were converted to phenotype codes (Phecodes) based on the PheWAS catalog~\cite{Den13}. ECGs with missing metadata were excluded. When a patient had multiple ECGs during a single admission, we used only the first ECG of their admission. 

\subsection{Model and Prototype Inference}
\begin{figure}
    \centering
    \includegraphics[width=\linewidth]{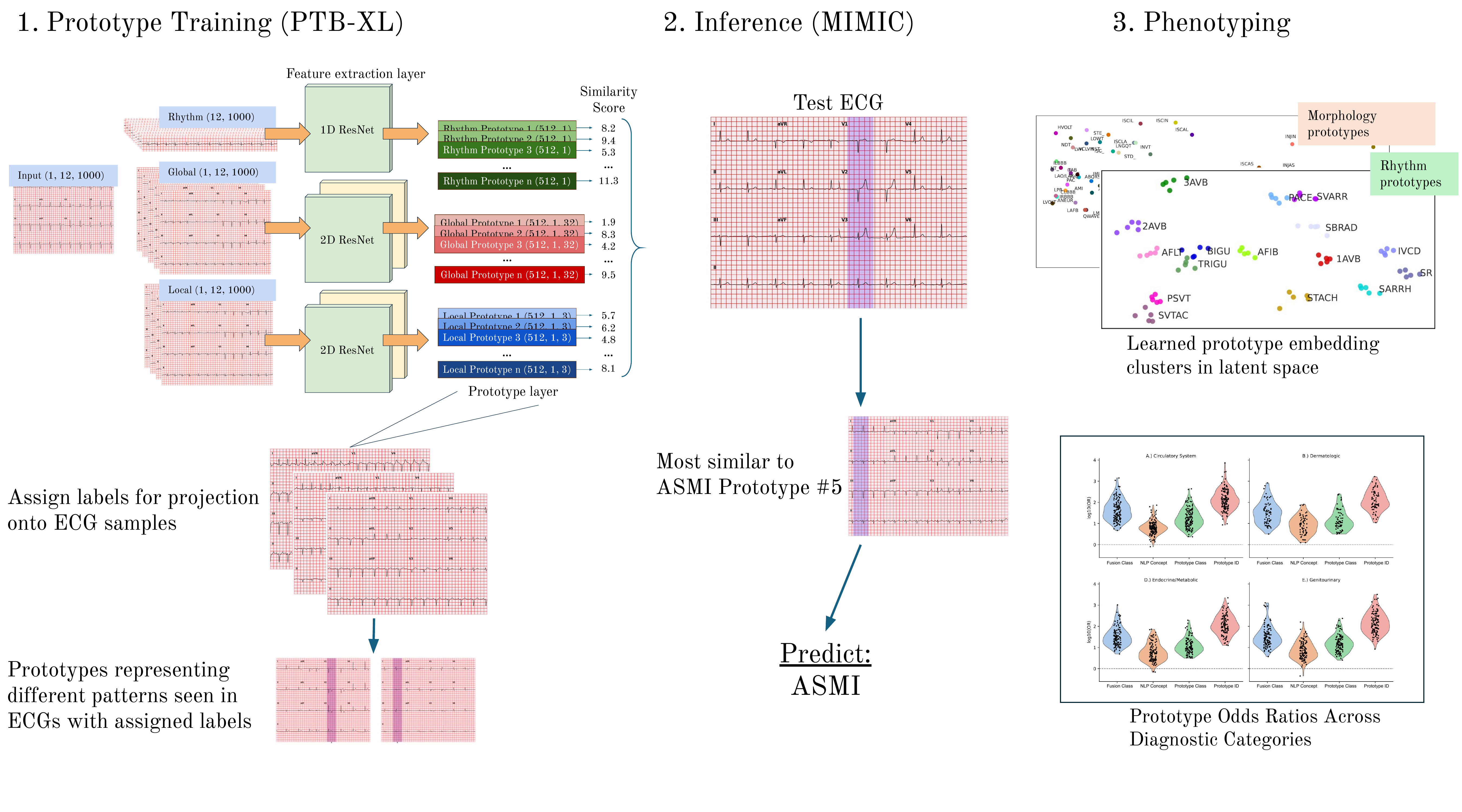}
    \caption{Overview of study approach. 1.) Prototype training: ProtoECGNet\cite{sethi_protoecgnet_2025} was trained on the PTB-XL dataset for multi-label ECG classification, with three branches (1D rhythm, 2D local morphology, 2D global) that each learned representative waveform prototypes. 2.) Inference: The pretrained model was applied without retraining to MIMIC-IV ECGs, computing similarity scores to identify the most representative prototypes for each recording. 3.) Phenotyping: Prototype activations and branch-level classes were associated with hospital discharge diagnoses (phecodes) and compared with NLP-extracted concepts, enabling statistical testing of whether prototypes capture clinically meaningful, transferable physiologic signatures.} 
    \label{fig:schematic}
\end{figure}

See Figure~\ref{fig:schematic} for an overview of our study approach. We used a prototype-based neural network architecture for multi-label ECG classification trained using methods previously described in the literature (code: \url{https://github.com/bbj-lab/protoecgnet})\cite{sethi_protoecgnet_2025}. In brief, the model architecture consisted of three parallel branches: a one-dimensional convolutional neural network (CNN) with global prototypes for rhythm detection, a two-dimensional CNN with local time-based prototypes for morphology recognition, and a two-dimensional CNN with global prototypes for whole-ECG patterns. Each branch was trained independently using binary cross-entropy loss and a prototype-based loss that promoted both clustering of within-class examples and separation of dissimilar ones. After training the prototype and feature extractor layers, each prototype was projected onto the latent space region most similar to it among training samples with its assigned label. Then, a final fully connected classifier was trained using the similarity scores from all prototypes across branches. During inference, the model was applied to both the PTB-XL test set and all eligible ECGs from MIMIC-IV. For each input ECG, we extracted the similarity score to each prototype and the most activated prototypes. No retraining or adaptation was performed on the MIMIC-IV data. As in Sethi et al. \cite{sethi_protoecgnet_2025}, the final model used five 1D rhythm prototypes, eighteen 2D partial prototypes, and seven 2D global prototypes per class.

\subsection{Prototype Grouping and Association Analysis}
\subsubsection{Prototype Analysis}
To discover higher-level phenotypic structure and reduce redundancy, we grouped prototypes based on similarity in the latent space. For each model branch, we computed pairwise cosine similarities between the final projected prototype vectors\cite{sethi_protoecgnet_2025}. This analysis visualized prototype embeddings across three ECG feature extraction approaches using principal component analysis (PCA). Prototype vectors were extracted from 1D rhythm branch features, 2D global features, and 2D morphology features, with each prototype assigned to a label held by its associated training ECG from PTB-XL. As described by Chen et al. \cite{chen_this_2019}, some prototypes may project onto the same training patch in the latent space and therefore be identical, but this is not an issue for classification tasks as the weights of the linear layer are tuned to accounted for this. Here, however, redundant prototypes were collapsed to facilitate analysis of independent prototypes. PCA dimensionality reduction was applied to transform high-dimensional prototype vectors into 2D coordinates for visualization. The resulting scatter plots displayed prototype distributions colored by clinical class, with centroid-based labeling used for the morphology features due to high density. This visualization approach enabled assessment of prototype clustering patterns and class separability across different feature extraction methodologies. 

\subsubsection{Extraction and Filtering of UMLS Concepts from ECG Reports}
To identify structured clinical concepts within unstructured electrocardiogram (ECG) reports, we used a natural language processing (NLP) pipeline based on the spaCy framework, augmented with biomedical extensions from SciSpaCy~\cite{Neu19}. Specifically, we employed the \verb|en_core_sci_scibert| language model, which is pretrained on large-scale biomedical corpora and optimized for clinical and scientific text. We augmented the pipeline with two components: (1) the \verb|scispacy_linker|, which performs entity linking to the Unified Medical Language System (UMLS)~\cite{Bod04}, and (2) the \verb|negex| component~\cite{Cha01} from \verb|negspacy|, which detects negated clinical concepts.

The \verb|scispacy_linker| was configured to resolve abbreviations and to prioritize entity mentions linked to definitions in UMLS, enhancing both precision and interpretability. For each report, the pipeline extracted named entities, resolved them to candidate UMLS concepts, and returned their concept unique identifiers (CUIs), canonical names, and linkage scores. Detected entities were also annotated for negation status based on syntactic context. Each report was processed individually, and the resulting concept-level annotations were aggregated into a structured list of UMLS concepts per report, including the original mention, concept metadata, and negation label.

To facilitate downstream analysis, we computed the frequency distribution of all extracted UMLS concepts across the corpus. Concepts which appeared at least 100 times were then manually reviewed and filtered to retain only those relevant to ECG interpretation—such as cardiac diagnoses (e.g., atrial fibrillation, myocardial infarction), waveform findings (e.g., ST elevation, QT prolongation), and procedural terminology (e.g., echocardiography references, pacemaker placement). This filtering step ensured that subsequent analyses focused on clinically meaningful ECG content rather than incidental mentions or non-cardiac concepts.

\subsubsection{Statistical Association with Clinical Diagnoses}
We performed association testing between the ECG-derived features and phenotypic outcomes using Fisher's exact tests on the full set of ECGs (using only the first ECG of an admission where multiple ECGs were performed). The categorical variables examined included ECG classification outcomes (“fusion”), prototype-based groupings across the three branches (1D, 2D partial, and 2D global prototypes), as well as the extracted binary medical concept indicators derived from UMLS CUIs. Phenotypic outcomes were represented by phecodes derived from the discharge diagnoses. Phecodes and CUIs present in at least 0.1\% of patients were included. For each phecode, 2×2 contingency tables were constructed comparing the presence/absence of each categorical feature value or CUI against phecode status. Multiple testing correction was applied using the Benjamini-Hochberg false discovery rate (FDR) method~\cite{Ben95}, with statistical significance defined as q $<$ 0.05.

\subsection{Prediction of Phecodes}
ECG data were preprocessed by parsing fusion class predictions and extracting best-matching prototypes for each of the 1D, 2D partial, and 2D global feature spaces using similarity scores to the patient's first ECG of their admission. Phecode outcomes were constructed as binary matrices from diagnostic codes, with prevalence filtering applied to retain phenotypes occurring in $\geq$0.1\% of patients. Four feature sets were evaluated: fusion classes, prototype classes, NLP-derived concepts (CUIs), and the combination of prototype embeddings and fusion classes (Prototype Combination). Logistic regression models were trained using subject-level train/test splits to prevent data leakage. Date shifting in MIMIC prevents temporal stratification of the dataset split. However, given the input features are an ECG rather than structured or clinically initiated data, there is unlikely to be significant dataset shift. Model performance assessed via area under the ROC curve (AUC) and 95\% confidence intervals computed through bootstrap resampling (n=1000). This framework enabled systematic comparison of different ECG feature representations for phenotype prediction across multiple cardiovascular and systemic conditions. The conditions of interest were chosen directly based on the Phecodes examined in Hughes et al\cite{hughes_deep_2025}. 

\subsection{Software}
 All data processing and statistical analyses were conducted using Python (NumPy, SciPy, scikit-learn) and PostgreSQL for database querying. ECG visualization and clustering analyses were performed using matplotlib and seaborn. ProtoECGNet\cite{sethi_protoecgnet_2025} was trained using a Nvidia A100 (40GB PCIe) GPU with 3.0 GHz AMD Milan processors in a HIPAA-compliant environment. Computation for this work used AMD Milan processors and took less than 24 hours to run inference on the MIMIC-IV dataset and perform association testing.

\section{Results}
\begin{figure}
    \centering
    \includegraphics[width=\linewidth]{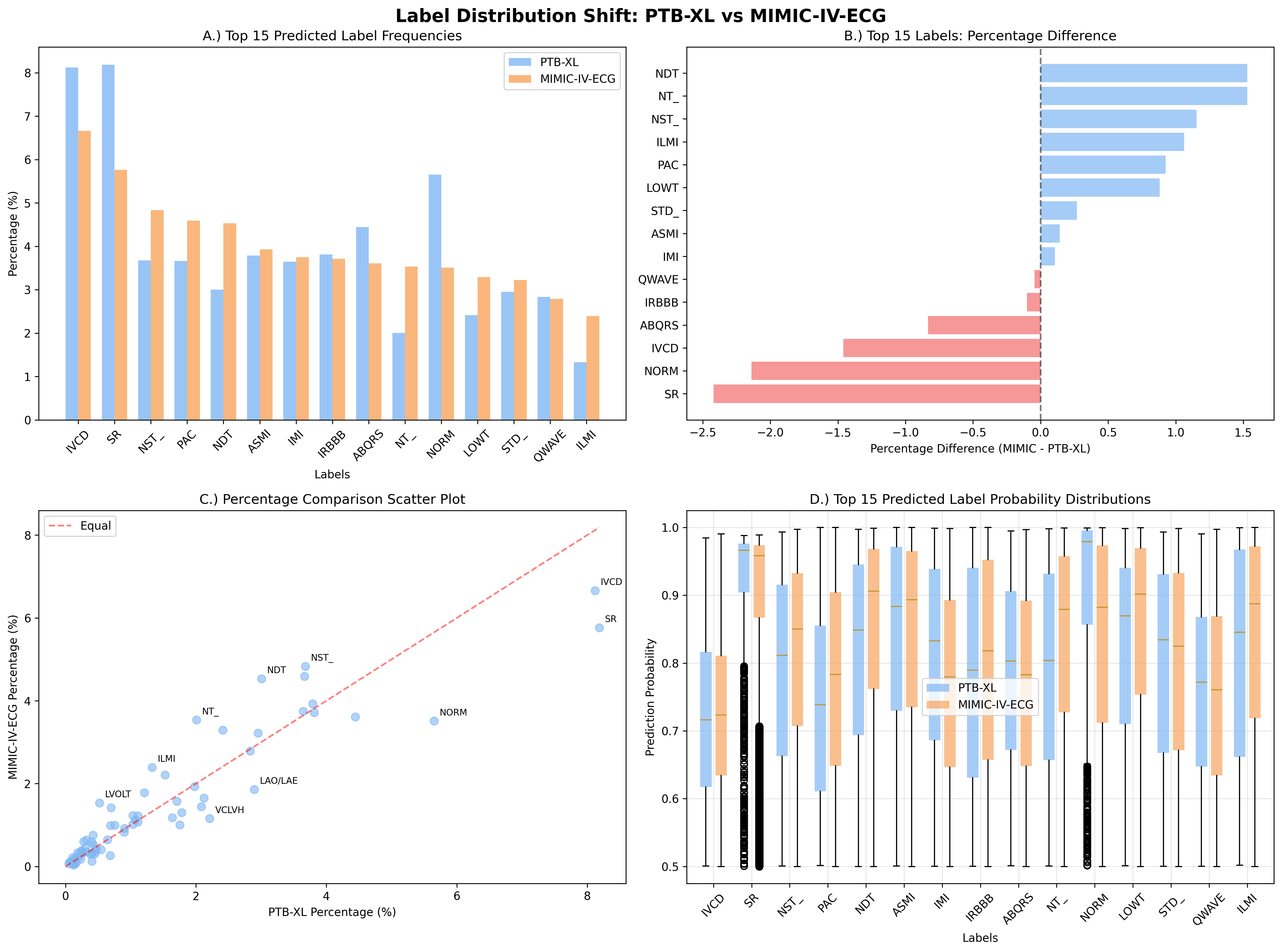}
    \caption{
        Comparison of predicted label distributions between PTB-XL and MIMIC-IV-ECG. \textbf{A.)} Frequencies (percentage of samples with predicted probability $\geq$ 0.5) of the 15 most prevalent predicted labels in MIMIC-IV-ECG, compared across both datasets. \textbf{B.)} Percentage difference in predicted label prevalence between MIMIC-IV-ECG and PTB-XL. Positive differences indicate higher predicted prevalence in MIMIC-IV-ECG; negative differences indicate lower prevalence compared to PTB-XL. \textbf{C.)} Scatterplot comparing label frequencies in PTB-XL and MIMIC-IV-ECG, illustrating broad agreement in prevalence patterns despite dataset differences (e.g., ICU setting in MIMIC). Labels with an absolute percentage difference greater than 1.0\% are annotated. The dashed line indicates equal prevalence across datasets. \textbf{D.)} Distribution of predicted probabilities for the 15 most prevalent labels in MIMIC-IV-ECG, shown for both PTB-XL and MIMIC-IV-ECG among samples where that label was predicted with probability $\geq$ 0.5.
    }
    \label{fig:label_distribution}
\end{figure}

This work evaluates whether the interpretable prototypes learned in Sethi et al.~\cite{sethi_protoecgnet_2025} are potentially useful intermediate or digital biomarkers. To do this, we took the ProtoECGNet model trained on the labeled PTB-XL dataset and performed inference on ECGs from MIMIC-IV-ECG \cite{johnson_mimic-iv_2023}. Within MIMIC there are computer generated reports, but the cardiologists' interpretations are not available—discharge summaries are the only available notes. This means there are no gold-standard labels for ECG class prediction. Thus, in Figure~\ref{fig:label_distribution}, we show ProtoECGNet's predicted classes using PTB-XL and MIMIC to compare the populations. While many labels are predicted at near equal ratios between the two datasets, there are several predicted substantially less frequently in MIMIC, particularly NORM (normal ECG) and SR (sinus rhythm)—likely reflecting the higher patient acuity in the MIMIC cohort, which consists solely of patients admitted to the hospital. 

\begin{figure}
    \centering
    \includegraphics[width=0.8\linewidth]{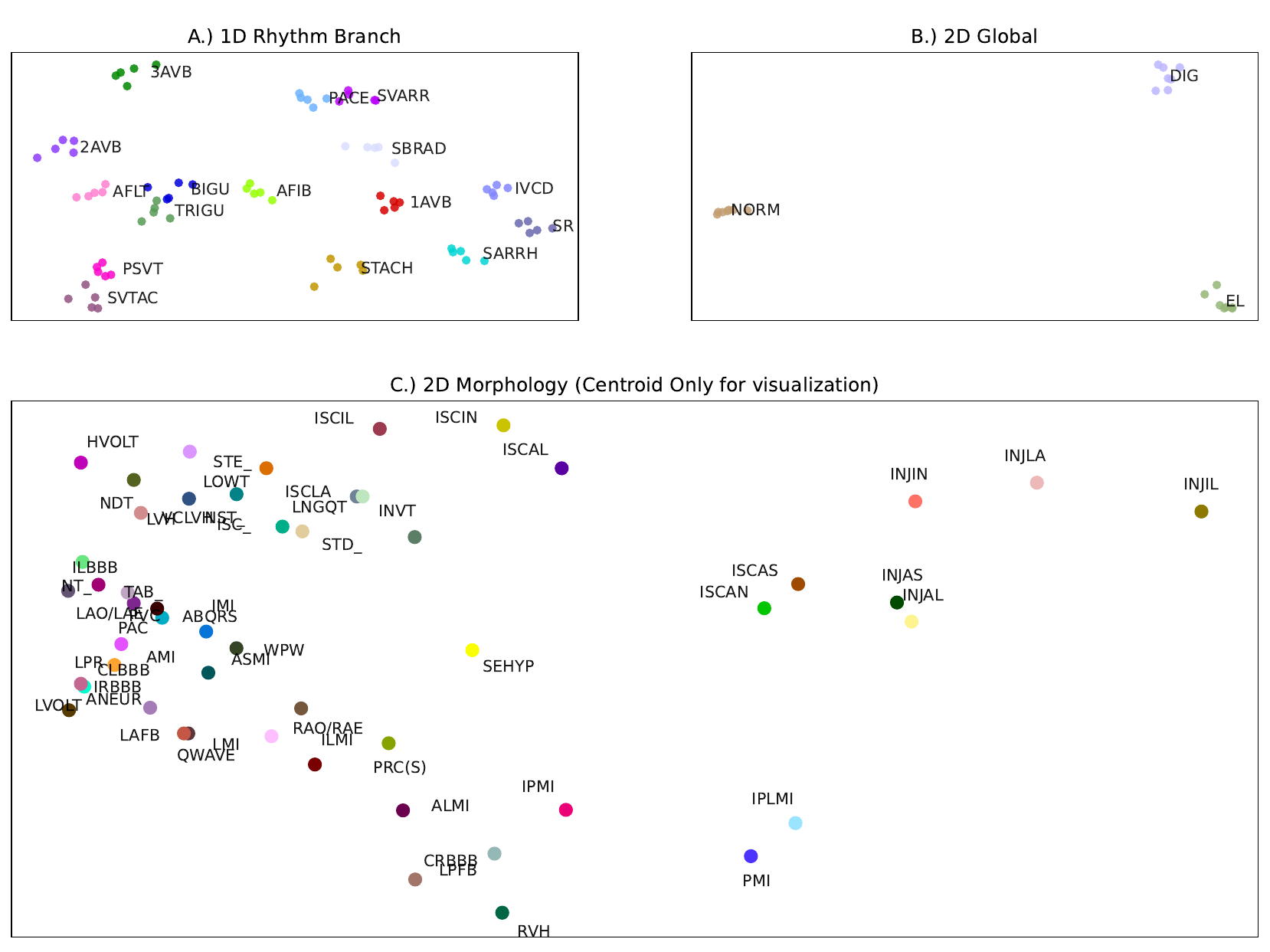}
    \caption{Principal Component Analysis of the prototype vector embeddings for each class within the 3 branches. A.) Shows each of the 5 prototypes for the 1D rhythm branch, B.) the 7 prototypes for each class in the 2D global branch, and C.) centroids for the 18 prototypes per class in the 2D morphology branch. Only centroids are shown to reduce over-plotting and to allow for visualization. See Sethi et al. for the full list of class abbreviation definitions. \cite{sethi_protoecgnet_2025}}
    \label{fig:PrototypePCA}
\end{figure}

Our prior work found advantages to learning prototypes across three branches, reflecting the way a human clinician might read an ECG: a 1D CNN with global prototypes for rhythm classification, a 2D CNN with time-localized prototypes for morphology-based reasoning, and a 2D CNN with global prototypes for diffuse abnormalities\cite{sethi_protoecgnet_2025}. To better understand the label space, we performed principal component analysis (PCA) to the vectors of the learned prototypes from ProtoECGNet within each class (Figure~\ref{fig:PrototypePCA}).

In the rhythm branch, the two principal components aligned with core interpretive axes used by clinicians: the vertical axis appeared to reflect heart rate, separating bradyarrhythmias (e.g., third-degree AV block) from tachyarrhythmias (e.g., supraventricular tachycardia), while the horizontal axis partially separated sinus from non-sinus rhythms. Sinus rhythms (SR [sinus rhythm], SARRH [sinus arrhythmia], STACH [sinus tachycardia], SBRAD [sinus bradycardia]) formed a compact cluster, with 1AVB (first-degree atrioventricular block) positioned slightly superior to SR, consistent with its more frequent occurrence in low heart rate contexts. Atrial arrhythmias (AFIB [atrial fibrillation], AFLT [atrial flutter]) appeared near one another despite differing morphology. Bigeminy (BIGU) and trigeminy (TRIGU) appropriately clustered tightly, as did the supraventricular tachycardias SVTAC (supraventricular tachycardia) and PSVT (paroxysmal supraventricular tachycardia). One anomaly was SVARR (supraventricular arrythmia), which clustered with pacemaker rhythms—potentially reflecting latent similarity to atrially paced morphologies.

In the 2D morphology branch, PCA revealed structure consistent with anatomical and electrical localization. Posterior myocardial infarctions (PMI [posterior myocardial infarction], IPLMI [inferoposterolateral myocardial infarction], IPMI [inferoposterior myocardial infarction]) clustered with right-sided conduction abnormalities (CRBBB [complete right bundle branch block], LPFB [left posterior fascicular block], RVH [right ventricular hypertrophy]), reflecting shared manifestations in right precordial leads. In contrast, anterior (AMI) and inferior (IMI) myocardial infarctions clustered separately. Ischemic syndromes (ISC [nonspecific ischemia], ISCLA [ischemia in lateral leads], ISCIN [ischemia in inferior leads], ISCAL [ischemia in anterolateral leads], ISCIL [ischemia in inferolateral leads]) formed a cohesive group, along with ST depressions (STD) and inverted T waves (INVT), which are often diagnostic features of ischemia. Hypertrophy patterns were more dispersed but showed partial alignment with other ECG patterns: RVH (right ventricular hypertrophy) was near right-sided conduction abnormalities, while LVH (left ventricular hypertrophy) appeared near left-sided conduction abnormalities, HVOLT (high QRS voltage), and nonspecific T-wave abnormalities. Wide QRS diagnoses (e.g., ILBBB [incomplete left bundle branch block], CLBBB [complete left bundle branch block], ABQRS [abnormal QRS]) partially grouped, but not uniformly. The global 2D branch included only three labels (NORM [normal ECG], EL [electrolyte disturbance or drug effect], DIG [digitalis effect]), which were clearly separated in PCA space.

\begin{table}[t]
\begin{minipage}[t]{.48\textwidth}
    \centering
    \includegraphics[width=\linewidth]{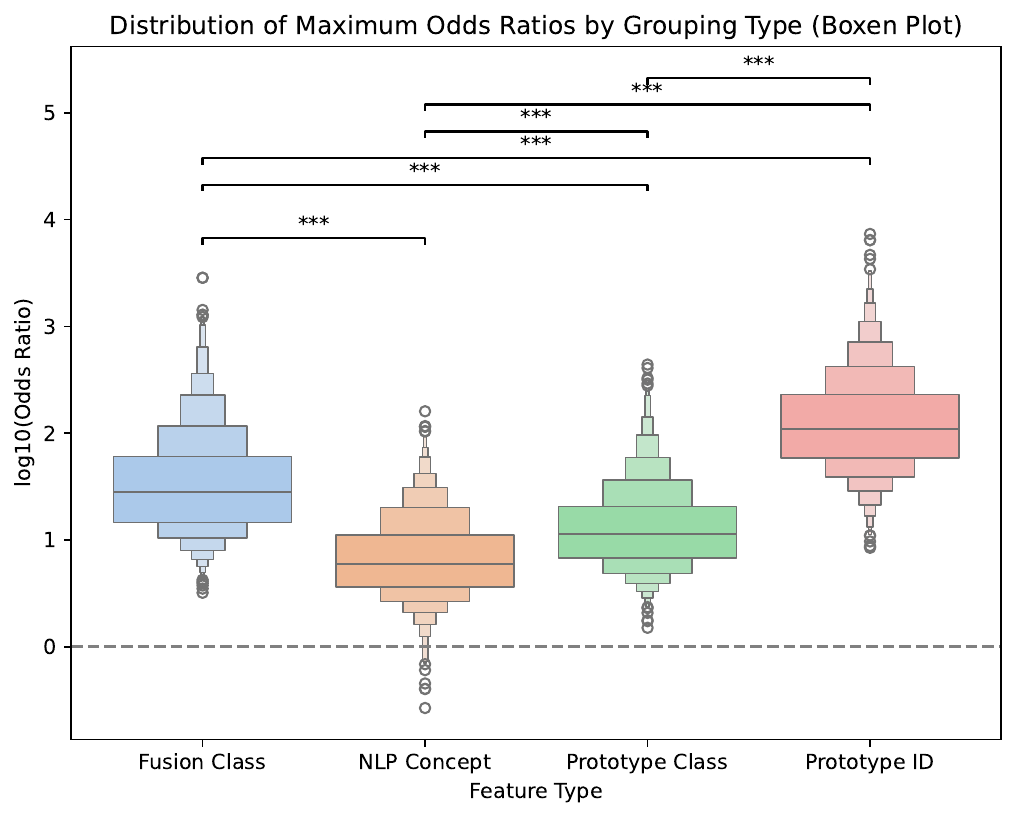}
\end{minipage}%
\hfill%
\begin{minipage}[t]{.48\textwidth}
\centering
\vspace{-180pt}
\captionof{table}{Mean intra-class cosine distances for prototype classes with mixed or uniform significance status. A Mann–Whitney U test~\cite{Man47} showed a significant difference ($p < 0.0001$), and Spearman correlation between odds ratio and intra-class distance was $\rho = 0.14$, $p < 0.0001$.}
\vspace{0pt}
\resizebox{\textwidth}{!}{%
\begin{tabular}[t]{lrrrr}
\toprule
Status & N & Mean & Std & 95\% CI \\
\midrule
Mixed & 870 & 0.245 & 0.047 & [0.242, 0.248] \\
Uniform & 73750 & 0.228 & 0.056 & [0.228, 0.228] \\
\bottomrule
\end{tabular}
\label{tab:cosine_distance_stats}
}
\end{minipage}%
\captionof{figure}{(Left.) Comparison of odds ratios based on each class of available labels. Fusion Labels: ProtoECGNet final predictions, NLP Concept: extracted diagnosis related concepts from computer generated ECG reports, Prototype Class: Branch-level class predictions within ProtoECGNet, and Prototype ID: branch-level individual prototypes most similar to a particular ECG. All pairwise comparisons are significant (*** = $p$-value $<$ 0.001). Prototype ID provides the cleanest groupings for associations to phenotypes.}
\label{fig:ORAll}
\end{table}


\begin{figure}
    \centering
    \includegraphics[width=0.88\linewidth]{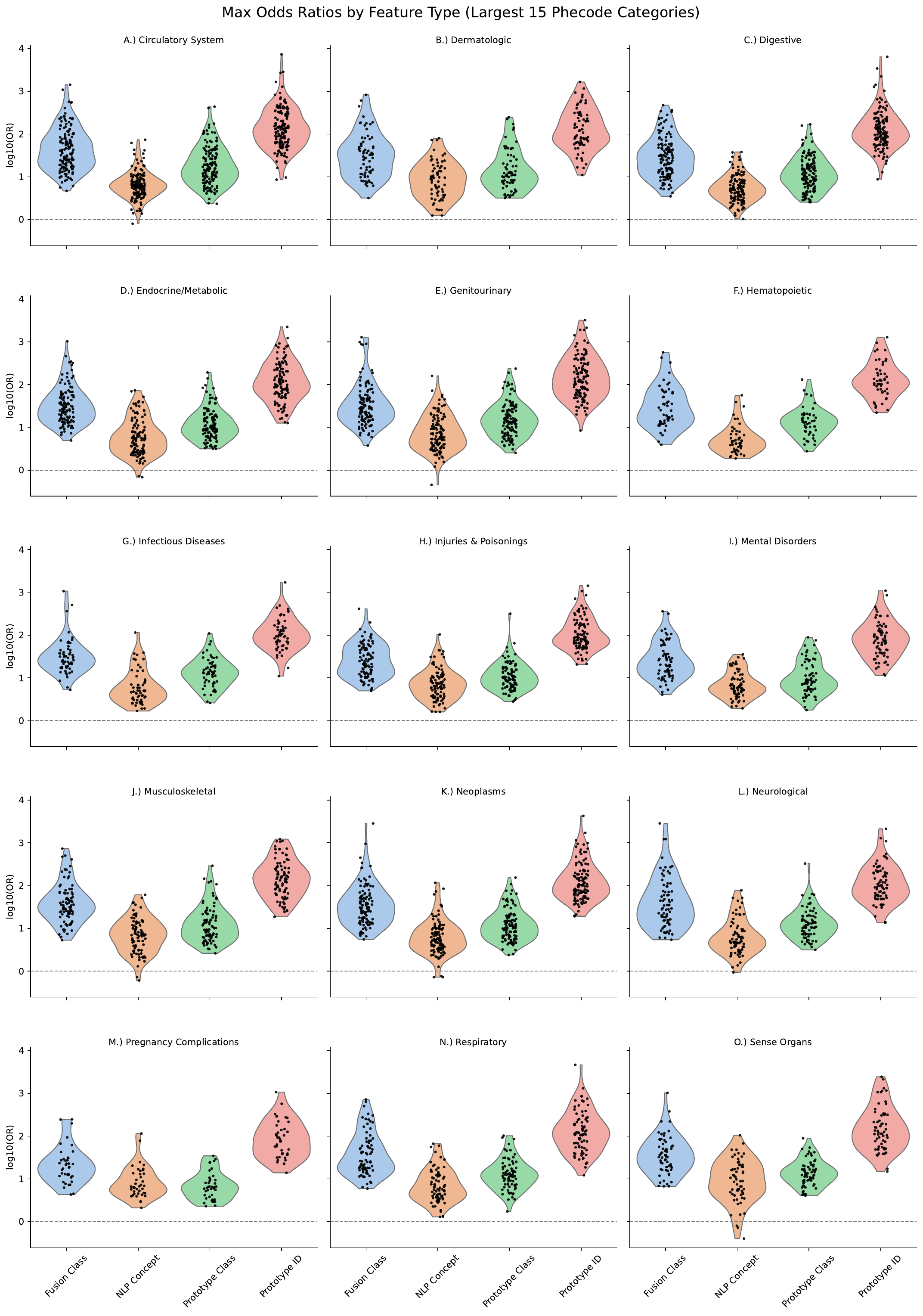}
    \caption{Comparison of odds ratio magnitude based on each set of available labels broken down across the 15 largest Phecode categories.}
    \label{fig:ORbyPhecode}
\end{figure}

Performing inference using the pretrained ProtoECGNet model produces labels at three potential levels: a.) fusion class labels are the final predicted classes based on all three branches, b.) prototype class labels are the class predictions within a single branch, and c.) prototype ID labels are the most representative prototypes for an example within the prototype class. An additional source of labels are the UMLS concepts which were extracted from the automated ECG interpretation currently in clinical use. We performed association testing via a PheWAS analysis using each of these four label groupings (Figure~\ref{fig:ORAll}). We observed both across all Phecodes (Figure~\ref{fig:ORAll}) and within major systems (Figure~\ref{fig:ORbyPhecode}) that performing the association at the individual prototype level yields stronger associations than any of the other modalities. This was even true within Phecodes categorized within the Circulatory System where there may be a tautological process between the computer-generated report and the discharge diagnoses.


As an additional verification of the learned prototypes, we compared the intra-class distances in the learned prototype embedding space (Table~\ref{tab:cosine_distance_stats}). We separated branch-level classes into groups: a.) Uniform: where all prototypes within a branch-level class were either significant or all were not significant based on the multiple testing corrected p-value, and b.) Mixed: where some of the prototypes were significant and others were not within the same branch-level class. We observe that the distance between the prototypes is significantly longer in classes where association significance was mixed (which indicates a more heterogeneous class). 

Finally, we evaluated whether the prototypes could be useful for the prediction of Phecodes at discharge in a similar manner to Hughes et al~\cite{hughes_deep_2025} in Table~\ref{t:results}. However, this cannot be a direct comparison due to two major differences. First, the models in Hughes et al. were explicitly trained to predict the outcome of interest from the raw ECG, and second, the sample size in MIMIC is substantially smaller. In contrast, ProtoECGNet was trained only to predict ECG labels. Despite these handicaps, we observed that using both the fusion classes and the representative prototypes (Prototype Combined) provided substantial signal and predicted many of the outcomes examined by Hughes et al. Intuitively, performance was especially strong for ECG-related diagnoses (e.g., atrial fibrillation), but we also observed similar trends to Hughes et al. for many of the other outcomes. This indicates that the prototypes learned by ProtoECGNet are potentially generalizable intermediate phenotypes or digital biomarkers which may be useful tools in disease specific and broader biological understanding. 

\begin{table}[ht]
\centering
\tbl{Performance of Phecode predictors based on input features (logistic regression).}{
\begin{scriptsize}
\begin{tabular}{l >{\raggedright\arraybackslash}p{4cm} rcccc}
\toprule
Phecode & Description & Training Cases & Fusion & Class & CUI & Prototype Combined \\
\midrule
38.0 & Septicemia & 3125 & 0.69 [0.68, 0.71] & 0.69 [0.67, 0.71] & 0.67 [0.65, 0.68] & 0.77 [0.76, 0.79] \\
38.1 & Gram negative septicemia & 1050 & 0.67 [0.64, 0.70] & 0.67 [0.63, 0.70] & 0.61 [0.58, 0.65] & 0.68 [0.65, 0.72] \\
260.0 & Protein-calorie malnutrition & 2215 & 0.61 [0.59, 0.63] & 0.58 [0.56, 0.61] & 0.58 [0.55, 0.60] & 0.67 [0.65, 0.70] \\
260.2 & severe protein-calorie malnutrition & 1055 & 0.60 [0.57, 0.64] & 0.60 [0.57, 0.64] & 0.61 [0.58, 0.65] & 0.71 [0.68, 0.75] \\
275.53 & Disorders of phosphorus metabolism & 1525 & 0.58 [0.55, 0.61] & 0.57 [0.54, 0.60] & 0.57 [0.53, 0.60] & 0.64 [0.60, 0.67] \\
276.11 & Hyperosmolality and/or hypernatremia & 2462 & 0.67 [0.65, 0.70] & 0.66 [0.64, 0.68] & 0.64 [0.62, 0.67] & 0.72 [0.70, 0.74] \\
276.6 & Fluid overload & 1328 & 0.62 [0.59, 0.65] & 0.62 [0.59, 0.65] & 0.56 [0.53, 0.59] & 0.67 [0.64, 0.70] \\
284.1 & Pancytopenia & 1217 & 0.54 [0.51, 0.58] & 0.54 [0.50, 0.57] & 0.57 [0.53, 0.60] & 0.59 [0.56, 0.62] \\
286.7 & Other and unspecified coagulation defects & 1112 & 0.65 [0.62, 0.69] & 0.65 [0.62, 0.68] & 0.63 [0.60, 0.66] & 0.73 [0.69, 0.76] \\
290.2 & Delirium due to conditions classified elsewhere & 2646 & 0.64 [0.62, 0.66] & 0.62 [0.60, 0.64] & 0.61 [0.59, 0.63] & 0.70 [0.68, 0.72] \\
317.11 & Alcoholic liver damage & 1747 & 0.64 [0.62, 0.67] & 0.64 [0.61, 0.67] & 0.64 [0.61, 0.67] & 0.70 [0.67, 0.73] \\
395.2 & Nonrheumatic aortic valve disorders & 3025 & 0.74 [0.72, 0.76] & 0.73 [0.72, 0.75] & 0.68 [0.66, 0.70] & 0.78 [0.76, 0.80] \\
395.6 & Heart valve replaced & 1792 & 0.77 [0.75, 0.79] & 0.78 [0.76, 0.80] & 0.74 [0.72, 0.76] & 0.86 [0.84, 0.88] \\
411.2 & Myocardial infarction & 9581 & 0.73 [0.72, 0.74] & 0.72 [0.70, 0.73] & 0.71 [0.70, 0.72] & 0.81 [0.79, 0.82] \\
411.8 & Other chronic ischemic heart disease, unspecified & 1360 & 0.83 [0.81, 0.85] & 0.81 [0.79, 0.83] & 0.79 [0.77, 0.82] & 0.90 [0.88, 0.92] \\
415.2 & Chronic pulmonary heart disease & 3250 & 0.72 [0.71, 0.74] & 0.71 [0.69, 0.73] & 0.70 [0.68, 0.72] & 0.80 [0.78, 0.82] \\
425.1 & Primary/intrinsic cardiomyopathies & 2034 & 0.79 [0.77, 0.81] & 0.77 [0.75, 0.80] & 0.77 [0.75, 0.79] & 0.88 [0.86, 0.90] \\
427.12 & Paroxysmal ventricular tachycardia & 1486 & 0.73 [0.70, 0.75] & 0.72 [0.69, 0.74] & 0.73 [0.70, 0.76] & 0.79 [0.76, 0.81] \\
427.21 & Atrial fibrillation & 15712 & 0.81 [0.80, 0.81] & 0.81 [0.80, 0.82] & 0.80 [0.79, 0.80] & 0.89 [0.88, 0.89] \\
427.22 & Atrial flutter & 1926 & 0.74 [0.72, 0.76] & 0.75 [0.72, 0.77] & 0.73 [0.70, 0.75] & 0.85 [0.83, 0.87] \\
427.42 & Cardiac arrest & 1212 & 0.71 [0.68, 0.74] & 0.69 [0.66, 0.72] & 0.70 [0.67, 0.73] & 0.73 [0.71, 0.76] \\
428.1 & Congestive heart failure (CHF) NOS & 8327 & 0.81 [0.80, 0.81] & 0.80 [0.79, 0.81] & 0.80 [0.79, 0.81] & 0.91 [0.90, 0.92] \\
428.3 & Heart failure with reduced EF [Systolic or combined heart failure] & 5807 & 0.83 [0.82, 0.84] & 0.82 [0.81, 0.84] & 0.80 [0.79, 0.81] & 0.88 [0.87, 0.89] \\
428.4 & Heart failure with preserved EF [Diastolic heart failure] & 5842 & 0.73 [0.72, 0.75] & 0.73 [0.72, 0.74] & 0.70 [0.69, 0.72] & 0.82 [0.80, 0.83] \\
501.0 & Pneumonitis due to inhalation of food or vomitus & 2325 & 0.67 [0.65, 0.69] & 0.65 [0.63, 0.67] & 0.64 [0.62, 0.66] & 0.72 [0.70, 0.74] \\
507.0 & Pleurisy; pleural effusion & 2746 & 0.65 [0.64, 0.68] & 0.63 [0.61, 0.66] & 0.62 [0.60, 0.64] & 0.72 [0.70, 0.74] \\
509.1 & Respiratory failure & 6355 & 0.67 [0.66, 0.69] & 0.67 [0.66, 0.68] & 0.65 [0.63, 0.66] & 0.77 [0.75, 0.78] \\
572.0 & Ascites (non malignant) & 1416 & 0.64 [0.61, 0.67] & 0.63 [0.60, 0.67] & 0.57 [0.54, 0.60] & 0.69 [0.66, 0.72] \\
585.32 & End stage renal disease & 2242 & 0.70 [0.68, 0.72] & 0.68 [0.66, 0.70] & 0.63 [0.60, 0.65] & 0.78 [0.76, 0.80] \\
994.2 & Sepsis & 4291 & 0.69 [0.68, 0.71] & 0.69 [0.68, 0.71] & 0.66 [0.65, 0.68] & 0.77 [0.76, 0.79] \\
\bottomrule
\end{tabular}
\end{scriptsize}
}
\label{t:results}
\end{table}

\section{Discussion}

We have demonstrated that prototype-based neural networks trained for ECG classification can capture clinically meaningful physiologic signatures or intermediate phenotypes that extend beyond their original training objectives. Our ProtoECGNet model, trained solely on the PTB-XL dataset for multi-label ECG classification, successfully transferred to the MIMIC-IV clinical database and revealed interpretable associations with a broad spectrum of clinical phenotypes, including both cardiac and non-cardiac conditions. 

Individual prototypes consistently showed stronger and more specific associations with clinical outcomes compared to fusion class predictions, NLP-extracted concepts, or broader prototype classes. This granular specificity suggests that the learned prototypes capture physiologically relevant waveform patterns that align with distinct pathophysiologic processes. Our analysis also revealed that prototype classes with mixed significance patterns (where some prototypes within a class were significantly associated with outcomes while others were not) exhibited significantly greater intra-class distances, indicating that the model learned to differentiate subtle but clinically meaningful variations within broader diagnostic categories. The prototypes also demonstrated substantial predictive utility across diverse conditions, achieving strong performance not only for cardiovascular phenotypes like atrial fibrillation (AUC 0.89) and heart failure (AUC 0.91), but also for conditions such as sepsis and renal disease, echoing the broad phenotypic associations previously observed in large-scale ECG studies.

Visualizing the learned prototype vectors revealed that the model's latent space structure mirrors key dimensions of clinical reasoning. In the rhythm branch, principal components aligned with heart rate and rhythm type—two foundational axes of ECG interpretation. Prototypes for sinus bradycardia, supraventricular tachycardia, atrial fibrillation, and paced rhythms were arranged along these axes in a manner that reflected their expected physiological and clinical relationships. Similarly, in the 2D morphology branch, prototypes for posterior myocardial infarctions clustered with right-sided conduction abnormalities, while ischemic syndromes grouped with ST depressions and T-wave inversions. These patterns suggest that, even without explicit supervision for phenotypic structure, the model organizes its internal representations to reflect established axes of clinical reasoning used in ECG interpretation.

Future work should focus on developing systematic approaches for optimal prototype selection and grouping~\cite{Bie11,Gar12}, potentially incorporating clinical expertise to ensure the most interpretable representations. Additionally, user studies involving cardiologists and trainees could evaluate how prototypes affect diagnostic accuracy and learning efficiency, potentially informing educational tools. Expanding this framework to incorporate temporal dynamics and multi-modal physiologic signals could also enhance the utility of learned phenotypes. Prospective validation would be essential to establish their generalizability and real-world utility.



A limitation of our analysis is that its retrospective nature limits causal inference about the relationships between ECG prototypes and clinical outcomes. Additionally, the absence of expert ECG annotations in MIMIC-IV prevented direct validation of our prototype interpretations against gold-standard clinical assessments. The smaller sample size in MIMIC-IV compared to previous large-scale ECG phenotyping studies may have also limited our power to detect associations with rarer conditions. Furthermore, potential dataset shift between the training (PTB-XL) and inference (MIMIC-IV) populations, including differences in patient acuity and clinical settings, may affect the generalizability of our findings. Finally, while our approach demonstrates the potential for interpretable digital phenotyping, the clinical implementation of such models would require careful consideration of regulatory requirements, workflow integration, and ongoing real-world clinical validation. Despite these limitations, this work demonstrates a promising direction to better incorporate physiological data into phenotyping pipelines even where task-specific gold-standard labels are not available.

\section{Acknowledgments}
This work was funded in part by the National Institutes of Health, specifically grant number R00NS114850 to BKB. Additionally, we would like to thank the University of Chicago Center for Research Informatics (CRI) High-Performance Computing team. The CRI is funded by the Biological Sciences Division at the University of Chicago with additional funding provided by the Institute for Translational Medicine, CTSA grant number UL1 TR000430 from the NIH.

\clearpage
\bibliographystyle{ws-procs11x85}
\bibliography{references,references-additional}

\end{document}